\documentclass[letterpaper, 10 pt, conference]{ieeeconf}  

\IEEEoverridecommandlockouts                              

\usepackage{graphicx}
\usepackage{moresize}
\usepackage[binary-units = true,
            quotient-mode=fraction,
            group-minimum-digits=5]{siunitx} 
\usepackage{afterpage}
\usepackage{threeparttable}
\usepackage{verbatim}
\usepackage{color}
\usepackage{tabu}
\usepackage{hyperref}
\usepackage{hyphenat}
\usepackage{color, colortbl}
\usepackage{multirow}
\usepackage{booktabs}
\DeclareSIUnit\px{px}
\DeclareSIUnit\batch{batch}
\DeclareSIUnit\image{image}
\definecolor{Blue}{rgb}{0.29,0.67,0.77}
\definecolor{Maroon}{rgb}{0.79,0.32,0.21}
\definecolor{Green}{rgb}{0.16,0.74,0.45}
\definecolor{Red}{rgb}{0.94,0.07,0.07}

\title{\LARGE \bf
Monocular Fisheye Camera Depth Estimation \\
Using Sparse LiDAR Supervision
}

\author{Varun Ravi Kumar$^{1}$, Stefan Milz$^{1}$, Christian Witt$^{1}$, Martin Simon$^{1}$, Karl Amende$^{1}$, Johannes Petzold$^{1}$,  \\ Senthil Yogamani$^{2}$ and Timo Pech$^{3}$
\thanks{$^{1}$Valeo und Schalter und Sensoren GmbH, Driving Assistance Advanced Research, Kronach        {\tt\small varun-ravi.kumar@valeo.com}}%
\thanks{$^{2}$Senthil Yogamani is with Valeo Vision Systems, Ireland
        {\tt\small senthil.yogamani@valeo.com}}%
\thanks{$^{3}$Timo Pech is with Technische University, Chemnitz, Germany
        {\tt\small timo.pech@etit.tu-chemnitz.de}}%
}

\begin{document}
\maketitle\thispagestyle{empty}
\pagestyle{empty}

\begin{abstract}

Near\hyp field depth estimation around a self\hyp driving car is an important function that can be achieved by four wide\hyp angle fisheye cameras having a field of view of over 180$^{\circ}$. Depth estimation based on convolutional neural networks (CNNs) produce state of the art results, but progress is hindered because depth annotation cannot be obtained manually. Synthetic datasets are commonly used but they have limitations. For instance, they do not capture the extensive variability in the appearance of objects like vehicles present in real datasets. There is also a domain shift while performing inference on natural images illustrated by many attempts to handle the domain adaptation explicitly. In this work, we explore an alternate approach of training using sparse LiDAR data as ground truth for depth estimation for fisheye camera. We built our own dataset using our self\hyp driving car setup which has a 64\hyp beam Velodyne LiDAR and four wide angle fisheye cameras. To handle the difference in view\hyp points of LiDAR and fisheye camera, an occlusion resolution mechanism was implemented. We started with Eigen's multiscale convolutional network architecture~\cite{Eigen_15} and improved by modifying activation function and optimizer. We obtained promising results on our dataset with RMSE errors comparable to the state\hyp of\hyp the\hyp art results obtained on KITTI.
\end{abstract}

\section{Introduction} \label{intro}

Depth estimation from single camera images is an important basic task for self driving cars such as driver assistance systems to solve localization and perception problems. Predominantly, the challenge is an arduous process and it cannot be decoded directly from bottom\hyp up geometric cues. A single captured image scene may be congruous with infinite real world scenarios~\cite{Eigen_14}. Successful approaches have relied on structure from motion, shape\hyp from\hyp X, binocular and multi\hyp view stereo. These techniques hinge on the assumption of prior knowledge about the characteristic appearance and multiple observations of the scene of interest that are available. The aforementioned can occur via multiple viewpoints, layout and size of object needs, cues such as shading, or observations of the scene under different lighting conditions. To overcome this limitation, there has recently been a rise in the number of works that pose the task of single image depth estimation as a supervised learning problem~\cite{Eigen_15,Eigen_14,Liu_16}. These methods seek to directly predict the depth from a single RGB image for each pixel through deep learning models that have been modeled on large collections of ground truth depth data.

Humans excel at monocular depth estimation by exploiting cues such as motion parallax, linear perspective, shape from shading, relative size and occlusion~\cite{Howard_12}. Full scene understanding with our capability to precisely estimate depth appears to bolster from the combination of both top-down and bottom-up cues~\cite{Godard_16}.

For supervised deep learning a large amount of training data is required in order to achieve high accuracy and to generalize on new scenes. In indoor environments, RGB\hyp D cameras are used to generate ground truth depth data for this task. However, strong sunlight has an adverse effect on infrared interference and make depth information of those sensing devices extremely noisy. In outdoor applications, especially in the domain of self driving cars, LiDAR or other laser scanners\hyp are used to capture ground truth data. Since measurements from 3D lasers have usually a sparse nature, the depth variations are captured with less details than visible in the image. 

Additional to the use of real data, synthetic rendering of depth maps are used to generate ground truth data. Rendered images do not unveil the scene and fail to implement real image noise characteristics\hyp which are the two drawbacks of this method~\cite{Aachen}. Also, there is an inefficiency to generalize on new scenes by the model trained on this approach.

The motivation of this paper is to provide a baseline for single frame depth estimation based on sparse Velodyne data as ground truth for training. This paper builds upon the authors' previous work published in a short paper~\cite{depthvarun} and the contributions of this paper include:
\begin{enumerate}
\item Demonstration of a working prototype purely trained on sparse Velodyne LiDAR data. 
\item Demonstration of fisheye camera depth estimation using CNN.
\item Adapting training data to handle occlusion due to difference in camera and Velodyne LiDAR viewpoint.
\item Tailoring the loss function and training algorithm to handle sparse depth data.
\end{enumerate}

The rest of the paper is structured as follows. Section~\ref{related-work} provides a survey of convolution neural networks (CNN) based depth estimation. Section~\ref{proposed} discusses the details of the network architecture, loss function tailoring and training algorithms. Section~\ref{secresults} summarizes results on our internal fisheye camera dataset and provide a comparison with publicly available KITTI results. Finally, Section~\ref{conclusion} concludes the paper and provides potential future directions.

\section{Related Work} \label{related-work}

It has been noted that in recent years, several deep learning based approaches to monocular depth estimation are trained in a supervised way~\hyp{} which requires a single input image~\hyp{} with no assumptions about the scene geometry or types of objects which are present.
In monocular depth estimation only single images are used at the inference time. Saxena et al.~\cite{Saxena_09} pioneered the supervised\hyp learning based approach called Make3D patch-based model. The input images are initially over\hyp segmented into patches and the 3D location and orientation of local planes are estimated which illustrates each patch. Markov Random Fields are used to combine the monocular cues with the stereo correspondences. The drawback of planar based approximations including~\cite{hoiem_efros_hebert_2005} is realistic outputs can not be generated as they lack global context since the estimates are made locally. They can be hindered when it comes to modeling of thin structures.

Liu et al.~\cite{Liu_16} formulated an approach for depth estimation as a deep continuous Conditional Random Fields (CRF) learning problem. Instead of hand\hyp crafted features such as unary and pairwise terms, Liu used deep convolutional neural fields that permitted the CNN features of unary and pairwise potentials end\hyp to\hyp end for training by utilizing continuous depth and Gaussian assumptions on the pairwise potentials.

Ladicky et al.~\cite{Ladicky_14} improved the per pixel depth estimation to a lucid classifier estimating only the probability of a pixel present at an arbitrarily fixed canonical depth. After appropriate image transformations, the probability of any other depths can be achieved by implementing the same classifier. The vulnerability of independent approaches of depth estimation and semantic segmentation are aimed directly by improving and generalizing the overall approach. 

Karsch et al.~\cite{Karsch_12} recommended a k\hyp Nearest\hyp Neighbor (kNN) transfer mechanism which can achieve better alignment which hinges on SIFT Flow~\cite{liu_yuen_torralba_sivic_freeman_2008} to estimate depths from single images of static backgrounds. They accomplished better estimation with the scene of interest in videos with dynamic foreground coupled with augmentation of the latter with motion information. A major drawback of this approach is a requirement of a complete training dataset to be available at inference time. 

In the last few years, it has been observed that object classification and recognition~\cite{Alexnet,imonyanZ_14a,SzegedyLJSRAEVR_14} 
reap great success with the application of Convolutional Neural Networks. 
CNNs perform classification of a single or multiple object label for a complete input image and apply bounding boxes on a few objects in each scene of an image. In addition to this, a variety of tasks like 
pose estimation~\cite{TompsonJLB_14}, 
stereo depth~\cite{ZbontarL_14} 
and instance segmentation~\cite{GuptaGAM_14} incorporate CNNs. Most of these models use CNNs to find only local features, or generate descriptors of discrete proposal regions; in contrast, Eigen's network uses both local and global views to predict a variety of output types. 

Laina~\cite{LainaRBTN_16} illustrated that dense depth maps can be produced by using ResNet\hyp based encoder\hyp decoder architecture. Their approach is demonstrated to predict dense depth maps in indoor scenes using RGB\hyp  images for training. Through example images~\cite{karsch_liu_kang_2012,liu_salzmann_he_2014}
it is found that the idea of depth transfer can be used to predict depth map or integrate depth map prediction with semantic segmentation~\cite{Eigen_15,Ladicky_14,liu_gould_koller_2010}
in supervised training.

Single\hyp image based depth estimation has various hardware\hyp based solutions like performing depth from defocus using a modified camera aperture proposed by Levin et al.~\cite{levin_fergus_durand_freeman_2007} and the Kinect v2 uses time\hyp of\hyp flight and active stereo to record depth.

We have incorporated Eigen's~\cite{Eigen_15} core multi\hyp scale architecture to adapt to a single task of estimating depth with an output resolution twice the original. We could achieve similar qualitative results with a sparse dataset, obtained from Velodyne HDL\hyp 64L rotating 3D laser scanner with valid depth points ranging from 3k\hyp 25k after occlusion removal.

\section{Model Architecture} 
\label{proposed}

\begin{figure}[!tbp]
\centering
\includegraphics[width=0.5\textwidth, height=6cm,keepaspectratio]{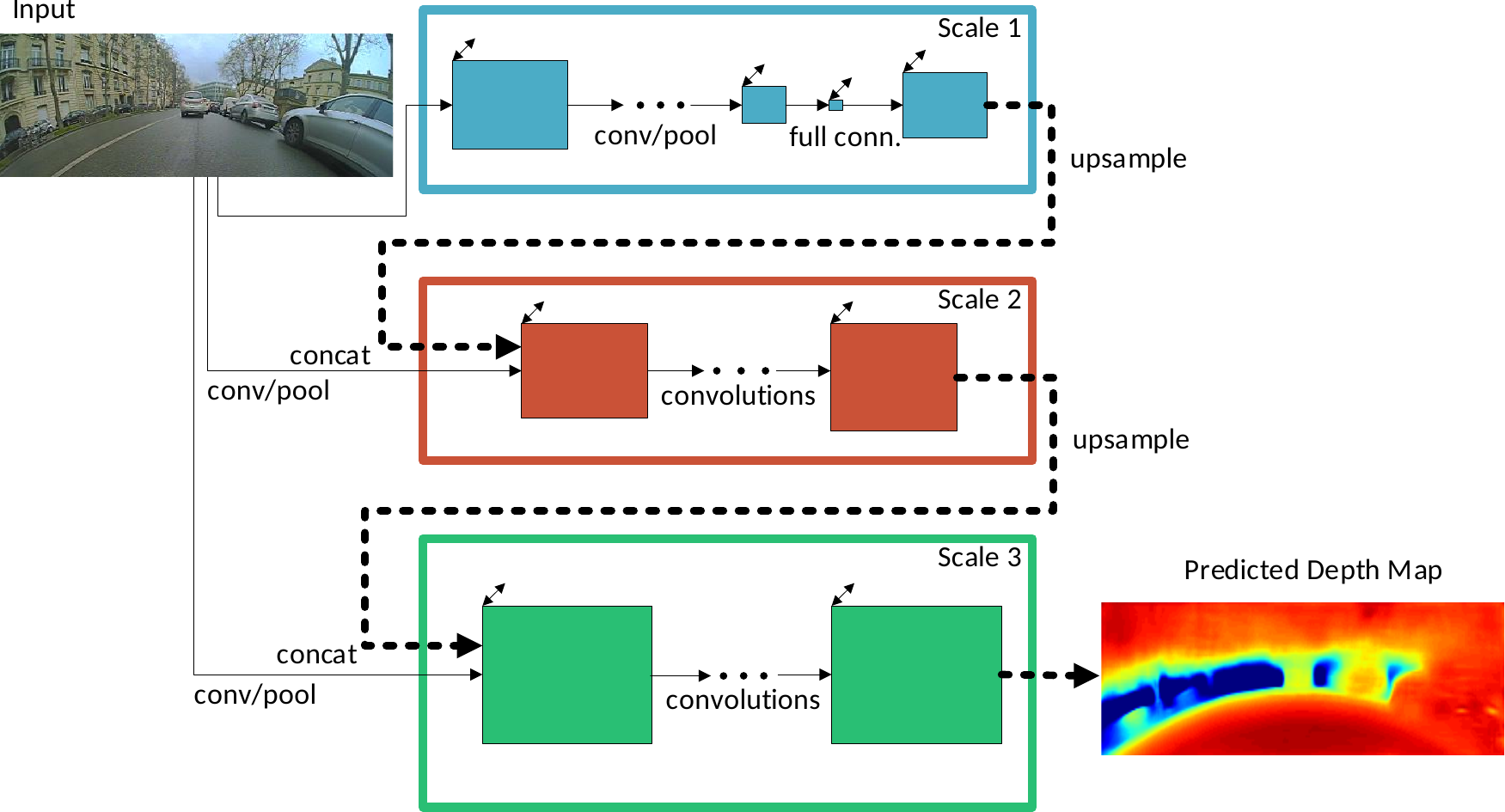}
\vspace{0pt}
\\
{
\ssmall
\setlength{\tabcolsep}{2pt}
\begin{tabular}{|l|l|ccccccc|c|}
\hline
 & Layer & 1.1 & 1.2 & 1.3 & 1.4 & 1.5 & 1.6 & 1.7 & upsamp \\
 \hline
\multirow{3}{*}{Scale 1}
 & Size   & 142x41 & 71x21 & 36x11 & 36x11  & 36x11 & 1x1 & 36x10 & 144x40 \\
\multirow{3}{*}{(AlexNet)}
 & \#convs & 1 & 1 & 1 & 1 & 1 & -- & -- & -- \\
 & \#chan & 96 & 256 & 384 & 384 & 256 & 4096 & 64 & 64 \\
 & ker. sz & 11x11      & 5x5   & 3x3   & 3x3   & 3x3 & -- & -- & -- \\
 & Ratio  & /8 & /16 & /16 & /16  & /32   & -- & /16 & /4\\
 & stride & 4 & 1 & 1 & 1 & 1 & -- & -- & \\
\hline
\hline
\multirow{5}{*}{Scale 2}
 & Layer & 2.1 & 2.2 & 2.3 & 2.4 & 2.5  &  &  & upsamp \\
 \hline
 & Size  & 284x82 & 142x40 & 142x40  & 142x40  & 142x40    && &   284x80 \\
 & \#chan & 96+64 & 64 & 64  & 64 & 1 & & & 1 \\
 & ker. sz  & 9x9 & 5x5   & 5x5   & 5x5   & 5x5 & & & -- \\
 & Ratio & /4 &  /4 & /4 &  /4 &  /4 &   &  & /2 \\
 & stride   & 2  & 1 & 1 & 1 & 1 & &  &  \\
\hline
\hline
\multirow{5}{*}{Scale 3}
 & Layer & 3.1 & 3.2 & 3.3 & 3.4 &  &  & & final\\
 \hline
 & Size  & 284x82 & 284x80  & 284x80  & 284x80  & & &  & 284x80  \\
 & \#chan & 64 & 64 & 64 & 1 & & & & 1 \\
 & ker. sz  & 9x9 & 5x5   & 5x5   & 5x5   &     & & & -- \\
 & Ratio & /2 & /2  & /2  & /2 &  &  &  &  /2 \\
 & stride   & 1  & 1 & 1 & 1 &  &  &  & \\
\hline
\end{tabular}
}
\caption{Multi\hyp{}scale architecture for depth prediction on raw fisheye images with a sparse velodyne (HD64L) ground truth. The input to the network is 576x172. Occlusion correction is essential, if velodyne points are mapped to the fisheye eye image plane, because of the different mounting positions of camera and LiDAR (see Section~\ref{Occlusion Correction}).}
\label{Multi-Scale architecture}
\end{figure}

Our model offers several architectural improvements to~\cite{Eigen_15} which is initially based on Eigen et al.~\cite{Eigen_14}. We adopted a simple architecture for Scale 1 based on AlexNet~\cite{Alexnet} to achieve real time on an embedded platform Nvidia TX 2. However, the usage of new model architectures such as ResNet\hyp 50~\cite{ResNet50} which have a bigger field of view could improve the results. These models take images of bigger dimensions as input and hence can provide a better global view of the image to the learning algorithm. Depending on the whole image area, a multi\hyp scale deep neural network first predicts a coarse global output and refines it using finer\hyp scale local networks.
This scheme is described in Fig.~\ref{Multi-Scale architecture}. 
The model is deeper with more convolutional layers compared to~\cite{Eigen_14}. Second, with the added third scale from~\cite{Eigen_15} at higher resolution, bringing the final output resolution up to half the input, or $\SI{284}{\px} \times \SI{80}{\px}$ for our sparse LiDAR fisheye camera dataset. In addition, we use swish~\cite{Swish} as the activation function rather than the mostly preferred rectified linear unit (ReLU)~\cite{ReLU}. Finally, we adopt Adam optimizer~\cite{Adam} which yields faster converging instead of the stochastic gradient descent (SGD) used by Eigen et al.~\cite{Eigen_15,Eigen_14}. Multi channel feature maps were passed similarly to~\cite{Eigen_15} avoiding the flow of output predictions from the coarse scale to the refine scale.

\paragraph{Scale 1: Full-Image View}
The first scale of the neural network 
analyses the global structure of the image and extracts global features. Global understanding of the scene requires an effective use of depth cues like object locations, vanishing points and alignment of structures~\cite{Eigen_15}. The local view of the image is inadequate to capture these features. Scale 1 is based on an ImageNet-trained AlexNet~\cite{Alexnet} with initialization of pre\hyp trained AlexNet weights only on convolutional layers. The global understanding of the image is achieved by two fully connected layers at the end. A very large field of view is obtained as each spatial location in the output connects to all the image features. The neural network takes fisheye images of size $\SI{576}{\px} \times \SI{172}{\px}$ as input. The output of the scale is a 64\hyp channel feature map with a resolution $\SI{142}{\px} \times \SI{40}{\px}$.

\paragraph{Scale 2: Predictions}
This scale incorporates a narrow view of the image and makes depth predictions at a resolution one\hyp fourth of the input image~\cite{Eigen_15}%
. While making predictions, the global scene information supplied by the Scale 1 is also considered by concatenation of feature maps. The input to this scale is the same RGB image which was given as input to Scale 1. 
Scale 2 corrects the coarse prediction it receives from Scale 1 to align with local details such as object and car edges, by concatenating the feature maps of the coarse network with those from a single layer of convolution and pooling. The output of the second scale is a $\SI{284}{\px} \times \SI{80}{\px}$ prediction for our sparse fisheye cameras dataset, with a single channel as a gray scale image.

\paragraph{Scale 3: Higher Resolution}
Scale 3 
refines the predictions made by Scale 2. It contains a set of convolutional operations with a small stride that can blend detailed structure of the image into the predictions. The alignment of output to higher\hyp resolution details is further refined which produces detailed spatially coherent depth map predictions. The final linear layer of this scale predicts the depth map with a resolution of $\SI{284}{\px} \times \SI{80}{\px}$.

\subsection{Sparse ground-truth depth maps}
\label{Sparsity}
A Velodyne HDL\hyp 64ES2 sensor can fire only 64 beams of lasers at different vertical angles with a vertical field of view of $\num26.8^{\circ}$. Hence the depth maps obtained from the projection of the LiDAR 3D points are sparse. Due to rotary motion of the Velodyne LiDAR sensor and movement of the vehicle while data recording was made, points that are far away had poor reflectivity. Therefore the extracted depth maps are sparser for scenes composed of far away objects.

\subsection{Scale-Invariant Error}
\label{Lossfunctionsection}
The sparse nature of the ground truth depth maps is considered in the design of the loss function.
We have adopted the loss function as described by Eigen et al.~\cite{Eigen_14} which is a $l_2$-loss with a scale\hyp invariant term. There is a lot of uncertainty regarding the global scale associated with the image, since we consider only a single image for depth prediction.
The scale\hyp invariant loss considers this scaling effect and produces the same loss for two 
scenes that differ only by the scaling factor. Last linear layer in the third scale of the architecture predicts the depth, which is compared to the ground truth depth map. The loss function is defined by equation \ref{loss_fn}, 

\begin{equation}
\label{loss_fn}
\mathrm{Loss}(p,p^*) =\frac{1}{n} \sum_{i \in V } d_{i}^2 - \frac{1}{n^2}\Bigg(\sum_{i \in V} d_{i}\Bigg)^2 
\end{equation}

where $p$ is the pixel wise set of predictions from the neural network. $p^*$ represents the ground truth depth map. Hence, $d_i = p_i-p_i^*$ is the difference for pixel $i$. The ground truth depth map is sparse, i.e. not for all pixels exists an equivalent depth measurement. We define a set of valid pixels $V \subset P^*$, with $V=\{p_1...p_i...p_n\}$, where $n$ is the number of valid pixels within the ground truth depth map~\cite{Eigen_14}:

\begin{equation}
\label{error_equation}
\mathrm{Loss}(p,p^*) = \frac{1}{n} \sum_{i \in V } (\log p_i - \log p_i^* + \alpha(p,p^*))^2.
\end{equation}

For a given ${(p,p^*)}$, the error is minimized by $\alpha$. The value of $\alpha$ is
${\alpha(p,p^*) = \frac{1}{n} \sum_{i \in V }(\log p_i^* - \log p_i)}$. The scale that best aligns to the ground truth is given by ${e^\alpha}$ for any prediction $p$. The error is same across all the scalar multiples of $p$, hence the term scale invariance as mentioned in~\cite{Eigen_14}.

An equivalent form of metric was obtained by Eigen et al.~\cite{Eigen_14} by setting ${d_i = \log p_i - \log p_i^*}$ to be the difference between the prediction and ground truth at pixel $i$,

\begin{equation}
\label{loss_fn_1}
\mathrm{Loss}(p,p^*) =\frac{1}{n^2} \sum_{i,j\in V } \big((\log p_i - \log p_j) - (\log p_i^* - \log p_j^*)\big)^2
\end{equation}

The error is demonstrated in equation~\ref{loss_fn_1} by comparing the relationships between pairs of pixels $i,j$ in the output: each pair of pixels in the prediction must differ in depth by an amount similar to that of the corresponding pair in the ground truth to have a low error.
Our fisheye dataset is extremely sparse due to the nature of LiDAR sensors, the loss function is adapted to this sparsity. By masking out pixels that do not have a valid depth value, the loss is calculated only on pixels which have depth values. This facilitates efficient feature extraction by the neural network. In addition to the scale\hyp invariant error, we 
evaluate our method using the error metrics used in~\cite{Eigen_14,Liu_16} as described in section~\ref{secresults}.
\subsection{Training\hyp Model}

We train our model in a single pass in an end\hyp to\hyp end fashion compared to~\cite{Eigen_15,Eigen_14} where the first two scales of the network were trained jointly. For each gradient step, the entire image area is considered for training. 
Pre\hyp trained weights from AlexNet~\cite{Alexnet} are used.
ConvNet is incorporated as a fixed feature extractor for our dataset and the last fully\hyp connected layers are removed. The fully connected layers are initialized randomly with values from a normal truncated distribution. Scale 2 and Scale 3 are randomly initialized. The dataset contains $\num{60000}$ images from fisheye camera and sparse Velodyne LiDAR scans as ground truth with validation and test set of $\num{5000}$ images each. We trained our model with a batch size of $\num{20}$ using the Adam~\cite{Adam} optimization algorithm, with $\beta_1 = 0.9$, $\beta_2 = 0.999$ and $\epsilon = 10^{-8}$. We adopt an exponential decay function to lower the learning rate as the training progresses, with an initial learning rate of $\lambda = 10^{-4}$. The function decays every $\num{7500}$ steps with a base of $\num{0.95}$. For the non\hyp linearities in the network, we used swish~\cite{Swish} activation function instead of the commonly used rectified linear units (ReLU)~\cite{ReLU} which tend to work better on deeper models. The swish function is defined as $f(x) = x \cdot \sigma(x)$ \cite{Swish},
where $\sigma(x) = ( 1 + e^{-x})^{-1}$ is the sigmoid function. 
The interesting aspect about the swish is that it does not monotonically increase compared to other activations functions like ReLU.
The problem of \textit{dead neurons} arises as the parameter will not be updated if the gradient is $0$, since gradient descent being the parameter update algorithm. We initially experimented by adopting different proposed alternative activation functions such as scaled exponential linear units (SELU)~\cite{SELU}, exponential linear units (ELU)~\cite{ELU} and leaky ReLU~\cite{Leaky_ReLU}. However, we found that swish performed best. 

\subsection{Occlusion Correction}
\label{Occlusion Correction}

\begin{figure}[!tbp]
 \centering
 \includegraphics[width=0.5\textwidth, height=10cm,keepaspectratio]{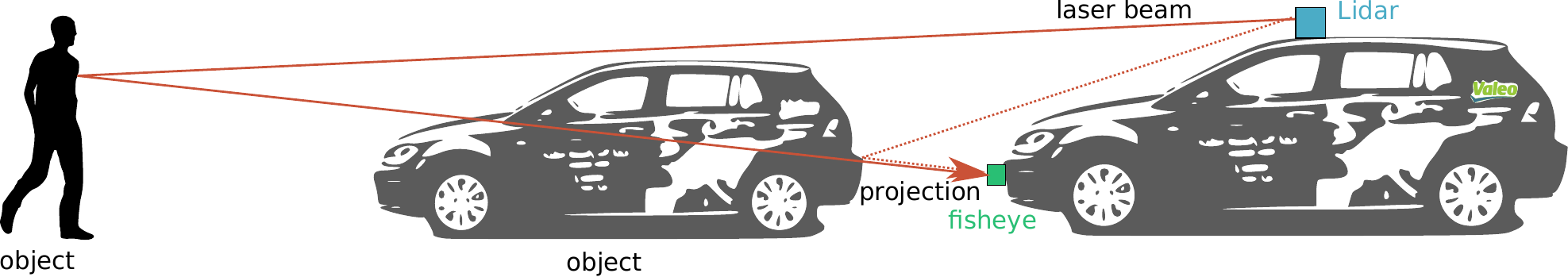}
    \caption{Illustraion of occlusion due to LiDAR's viewpoint being higher than the fisheye camera's viewpoint. 3D points from the object (person) will be mapped to image plane even though it is not visible from camera.}
    \label{Occlusion_Fig}
\end{figure}

The sensor fusion of the data will be correct, if both camera and the Velodyne LiDAR scanner beholds the world from the same viewpoint. However, for technical reasons in our vehicle the fisheye camera is in the front and the LiDAR is placed at the top as seen in Fig.~\ref{Occlusion_Fig}. LiDAR perceives the environment behind objects that occlude the view for the camera. This problem of occlusion results in wrong mapping of depth\hyp points that are not visible to the camera. It is hard to solve, since occluded points are projected adjacently to unoccluded points \cite{Disocclusion}.

\begin{figure}[!tbp]
 \centering
 \includegraphics[width=0.47\textwidth, height=5cm,keepaspectratio]{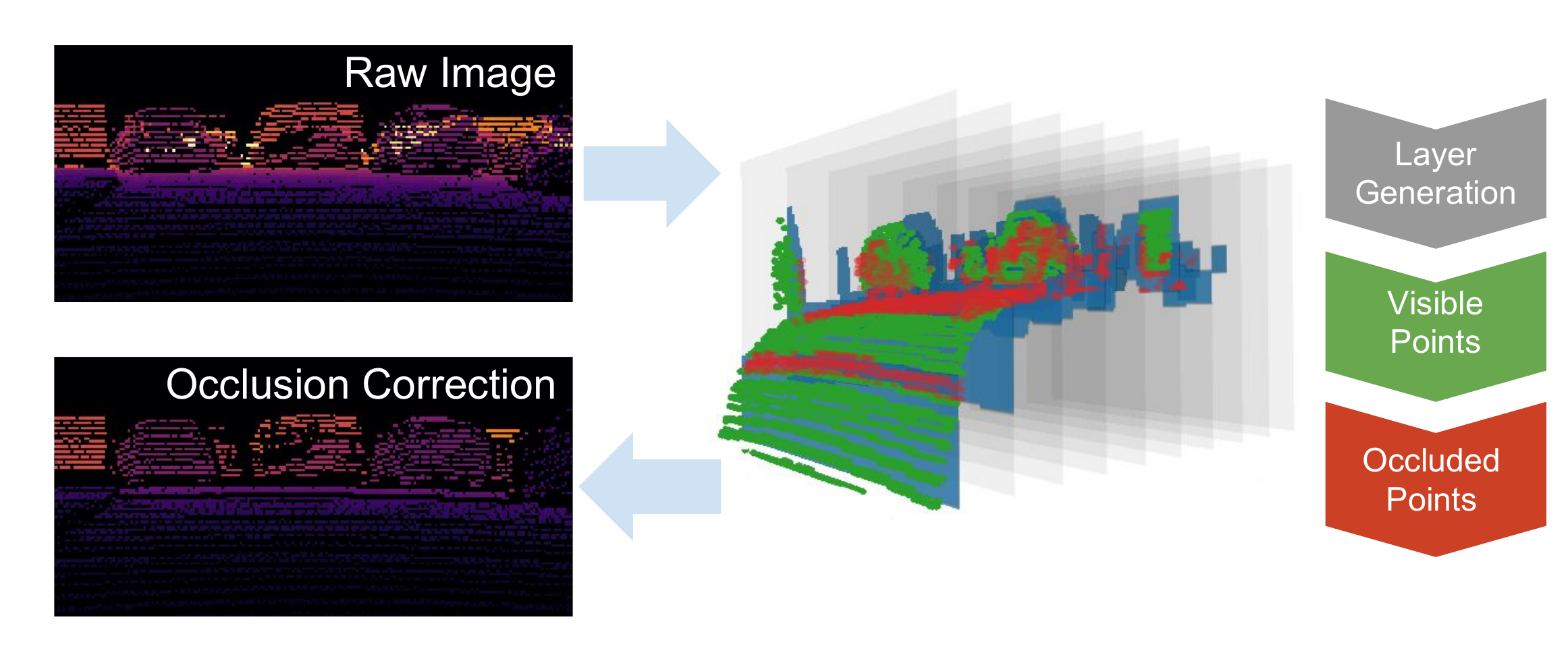}
    \caption{Visualization of the distance based segmentation technique with morphological filters.
    LiDAR points are projected to corresponding layers and are removed if
    they are occluded by dilated parts of a neighboring layer.}
    \label{Ocllusion_fig}
\end{figure}

To solve this problem, we adapted a distance based segmentation technique with morphological filters as shown in the Fig.~\ref{Ocllusion_fig}. Instead of directly projecting points from the LiDAR into the image plane of the fisheye camera, we introduce $I$ layers within the camera view located at a distance $d_i^{\text{layer}}$, $i = 1, \ldots, I$. Each LiDAR point will be projected onto the layer next to it.
We apply a morphological filter that dilates points within each layer to fill the sparse regions (in Fig.~\ref{Ocllusion_fig} dilated parts of the layers are colored blue).
A point at a distance $d^{\text{point}}$ is regarded as occluded, if a layer $i$ exists with $d_i^{\text{layer}}<d^{\text{point}}$.
Otherwise the valid point is projected onto the image plane of the fisheye camera. 

\begin{figure}[!tbp]
 \centering
 \includegraphics[width=0.47\textwidth, height=10cm,keepaspectratio]{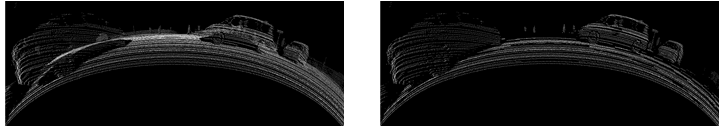}
    \caption{Illustration of Occluded Velodyne Ground Truth (left) and Dis-Occluded Velodyne Ground Truth (right)}
    \label{Occlucion_Correction}
\end{figure}

\begin{table*}[th]
\centering
\caption{Quantitative results of leaderboard algorithms on KITTI 2015~\cite{KITTIDepth} dataset and our approach on Valeo's fisheye dataset}
\label{results}
\resizebox{\textwidth}{!}{%
\begin{tabular}{lclccccccc}
\toprule
&  &  & RMSE & RMSE (log) & ARD & SRD & $\delta < 1.25$ & $\delta < 1.25^2$ & $\delta < 1.25^3$ \\
\cmidrule(lr){4-7} \cmidrule(lr){8-10}
Approach & Supervised & cap & \multicolumn{4}{c}{lower is better} & \multicolumn{3}{c}{higher is better}\\
\midrule
Mancini et al.~\cite{ManciniCVC16} 		          & Yes  & $0 - \SI{100}{\meter}$ & 7.508 & 0.524 & 0.196 & -     & 0.318 & 0.617 & 0.813 \\
Eigen et al.~\cite{Eigen_14} coarse 28$\times$144 & Yes  & $0 - \SI{80}{\meter}$  & 7.216 & 0.273 & 0.194 & 1.531 & 0.679 & 0.897 & 0.967 \\
Eigen et al.~\cite{Eigen_14} fine 27$\times$142   & Yes  & $0 - \SI{80}{\meter}$  & 7.156 & 0.270 & 0.190 & 1.515 & 0.692 & 0.899 & 0.967 \\
Liu et al.~\cite{LiuSL14} DCNF-FCSP FT            & Yes  & $0 - \SI{80}{\meter}$  & 6.986 & 0.289 & 0.217 & 1.841 & 0.647 & 0.882 & 0.961 \\
Ma et al.~\cite{fangchangma}       		          & Yes  & $0 - \SI{100}{\meter}$ & 6.266 & -    & 0.208  & -     & 0.591 & 0.900 & 0.962 \\
Kuznietsov et al.~\cite{Aachen}                   & Yes  & $0 - \SI{50}{\meter}$  & 3.531 & 0.183 & 0.117 & 0.597 & 0.861 & 0.964 & 0.989 \\
Zhou et al.~\cite{Zhou17} (w/o explainability)    & No   & $0 - \SI{50}{\meter}$  & 5.452 & 0.273 & 0.208 & 1.551 & 0.695 & 0.900 & 0.964 \\ 
Zhou et al.~\cite{Zhou17}                         & No   & $0 - \SI{50}{\meter}$  & 5.181 & 0.264 & 0.201 & 1.391 & 0.696 & 0.900 & 0.966 \\
Godard et al.~\cite{Godard_16}                    & No   & $0 - \SI{50}{\meter}$  & 4.471 & 0.232 & 0.140 & 0.976 & 0.818 & 0.931 & 0.969 \\
\midrule\midrule
Ours fine 80$\times$284                  & Yes &  $0 - \SI{50}{\meter}$ & 1.717 & 0.236 & 0.160 & 0.397 & 0.816 &0.934 & 0.969 \\
\bottomrule
\end{tabular}%
}
\end{table*}

\section{Results} \label{secresults}
The model is completely trained on our internal dataset. Our dataset contains $\num{55000}$ images obtained from raw fisheye camera and sparse Velodyne HDL\hyp 64E rotating 3D laser scanner as ground truth. Points without depth value are left unfilled without any post\hyp processing. Eigen's model~\cite{Eigen_15} handles missing values by eliminating them in the loss function. 
The input 
images are down\hyp sampled 
to $\SI{576}{\px} \times \SI{172}{\px}$ primarily to get faster inference and training times.

The ground truth depth for this dataset is captured at various intervals using a Velodyne HDL\hyp 64E rotating 3D laser scanner, and are sampled at irregularly spaced points. Conflicting values are found when constructing the ground truth depths for training, since sensor records data at a set maximum frequency of $\SI{10}{\hertz}$ and the fisheye cameras record data at $\SI{30}{\hertz}$. Time synchronization is essential as the sensors capture data at different frequencies. Each spin of the LiDAR sensor is considered as a frame and carries a time\hyp stamp associated with it. Similarly, each image frame recorded by the fisheye camera carries a time\hyp stamp. For the purpose of synchronization, time\hyp stamps provided with the recordings are used. We resolve conflicts by choosing the depth recorded closest to the RGB capture time in Intempora RTMaps (Real\hyp Time Multisensor Applications) framework.

The training set was collected by driving around Paris, France and various parts of Bavaria, Germany. The training set includes scenes from the \textit{city}, \textit{residential} and \textit{sub\hyp urban} categories of our raw dataset. These are randomly shuffled and fed to the network. We train the entire model for $\num{80}$ epochs and test prediction takes $\SI{3.45}{\second/\batch}$ with a batch size of 20 images ($\SI{0.17}{\second/\image}$).

\begin{figure*}[!tbp]
  \centering
  \includegraphics[width=\textwidth, height=20cm, keepaspectratio]   {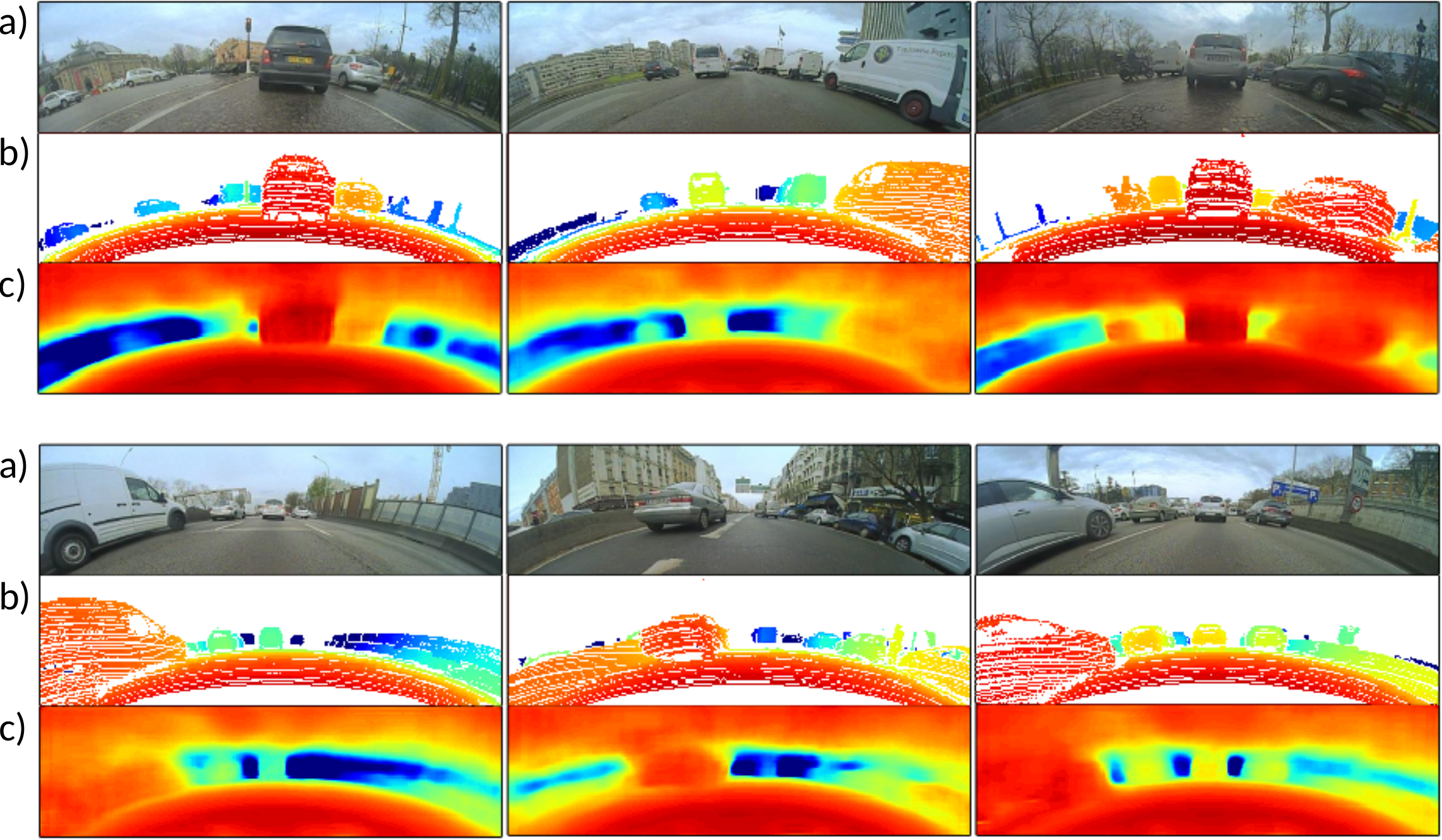}
  \caption{Qualitative results: Exemplary predictions by the proposed CNN network. For each image, we show (a) RGB Input (b) LiDAR Ground Truth
(c) Predicted Depth Map [The sky is considered to be invalid pixel i.e masked as zero while training. We have not considered disparity depth for ground truth generation as compared to KITTI~\cite{KITTIDepth}. The depth values are in 8\hyp bit intensity range (0 - 255)]}
\label{examplePreds}
\end{figure*}

The evaluation of accuracy in our method in depth prediction is using the 3D laser ground truth on the test images. We use the depth evaluation metrics used by Eigen et al.~\cite{Eigen_14}. Exemplary predictions are shown in figure \ref{examplePreds}. The qualitative results show that image regions without sufficient large ground truth data points (e.g. sky), the model fails to predict reasonable values. 

A protocol evaluation is applied and results are shown by discarding ground-truth depth below $\SI{0}{\meter}$ and above $\SI{50}{\meter}$ while capping the predicted depths into $\SI{0}{\meter} - \SI{50}{\meter}$ depth interval. This implies, we set predicted depths to $\SI{0}{\meter}$ and $\SI{50}{\meter}$ if they are below $\SI{0}{\meter}$ or above $\SI{50}{\meter}$, respectively.

In Table \ref{results}, we show how our approach performs on Valeo's fisheye dataset. 
Furthermore the results of leaderboard algorithms on KITTI 2015~\cite{KITTIDepth} are reproduced. For lack of a better comparison, we use this as a proxy to illustrate that we obtained comparable RMSE on our sparse fisheye dataset. 
It should be noted that although we predict a dense depth map, the sparse dataset only allows us to take a fraction of the predicted values into consideration for error calculation. To tackle this problem we plan to refine our model on a synthetic dataset, close to our Valeo's fisheye dataset, that allows a full verification of the predicted depth. First tests show promising results with excluded sky.

\section{Conclusion} \label{conclusion}

Even though the camera/LiDAR setups are different, the results provide a reasonable comparison to KITTI on performance of monocular depth regression using sparse LiDAR input. In future work, we aim to improve the results by using more consecutive frames which can exploit the motion parallax and better CNN encoders. We also plan to augment the supervised training with synthetic data and unsupervised training techniques.
\bibliographystyle{IEEEtran}
\bibliography{egbib}
\end{document}